\documentclass[letterpaper, 10 pt, conference]{ieeeconf}  

\IEEEoverridecommandlockouts                              

\usepackage{microtype}
\usepackage{hyperref}

\usepackage{varwidth}

\usepackage{graphicx}
\usepackage{booktabs} 
\usepackage{amssymb} 
\usepackage{amsmath}
\usepackage{mathtools}
\usepackage{multirow}
\usepackage{bm}
\usepackage{upgreek}
\usepackage{algorithmic}
\usepackage{algorithm}
\usepackage{tabularx}
\usepackage{graphicx}
\usepackage{subfig}
\usepackage{tikz}
\usepackage{adjustbox}
\usetikzlibrary{shapes.geometric, arrows}
\usepackage{color, xcolor}
\usepackage{soul}
\usepackage[normalem]{ulem}
\tikzstyle{block} = [rectangle, draw, fill=blue!20, text width=5em, text centered, rounded corners, minimum height=2em]
\tikzstyle{line} = [draw, -latex']

\title{\LARGE \bf
Dual-Arm Hierarchical Planning for Laboratory Automation: Vibratory Sieve Shaker Operations
}

\author{Haoran Xiao$^{1}$, Xue Wang$^{1}$, Huimin Lu$^{1}$, Zhiwen Zeng$^{1}$, Zirui Guo$^{1}$, Ziqi Ni$^{2}$, Yicong Ye$^{2}$, Wei Dai$^{1}$
\thanks{$^{1}$College of Intelligence Science and Technology, and the National Key Laboratory of Equipment State Sensing and Smart Support, National University of Defense Technology, Changsha, China.}
\thanks{$^{2}$College of Aerospace Science and Engineering, National University of Defense Technology, Changsha, China.}%
\thanks{Haoran Xiao and Xue Wang contribute equally to this work.}%
\thanks{This work was supported in part by the National Science Foundation of China under Grant 62203460, U22A2059, and the Innovation Science Foundation of National University of Defense Technology under Grant 24-ZZCX-GZZ-11.}%
}
\begin{document}
\providecommand\BIBentryALTinterwordstretchfactor{2.5}
\maketitle
\thispagestyle{empty}
\pagestyle{empty}

\begin{abstract}
This paper addresses the challenges of automating vibratory sieve shaker operations in a materials laboratory, focusing on three critical tasks: 1) dual-arm lid manipulation in 3 cm clearance spaces, 2) bimanual handover in overlapping workspaces, and 3) obstructed powder sample container delivery with orientation constraints. These tasks present significant challenges, including inefficient sampling in narrow passages, the need for smooth trajectories to prevent spillage, and suboptimal paths generated by conventional methods. To overcome these challenges, we propose a hierarchical planning framework combining \textbf{Prior-Guided Path Planning} and \textbf{Multi-Step Trajectory Optimization}. The former uses a finite Gaussian mixture model to improve sampling efficiency in narrow passages, while the latter refines paths by shortening, simplifying, imposing joint constraints, and B-spline smoothing. Experimental results demonstrate the framework's effectiveness: planning time is reduced by up to 80.4\%, and waypoints are decreased by 89.4\%. Furthermore, the system completes the full vibratory sieve shaker operation workflow in a physical experiment, validating its practical applicability for complex laboratory automation.

\end{abstract}
\section{Introduction}
The operation of a vibratory sieve shaker (VSS) in materials laboratories presents a challenging automation scenario due to its unique requirements: precise manipulation in narrow passages, coordination between dual arms, and smooth trajectory execution to prevent material spillage. These challenges are exacerbated by the need to operate within tightly constrained workspaces and avoid collisions with both environmental obstacles and the robot's links.
\begin{figure}[!t]
\centering
\subfloat[]
  {\includegraphics[width=0.24\textwidth]{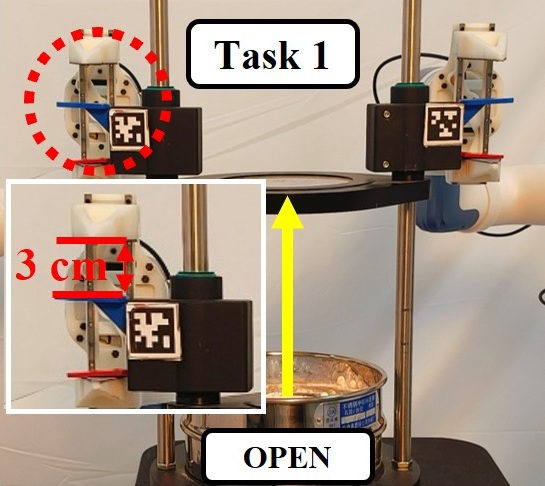}\label{fig1_a}} \hfill
\subfloat[]
  {\includegraphics[width=0.24\textwidth]{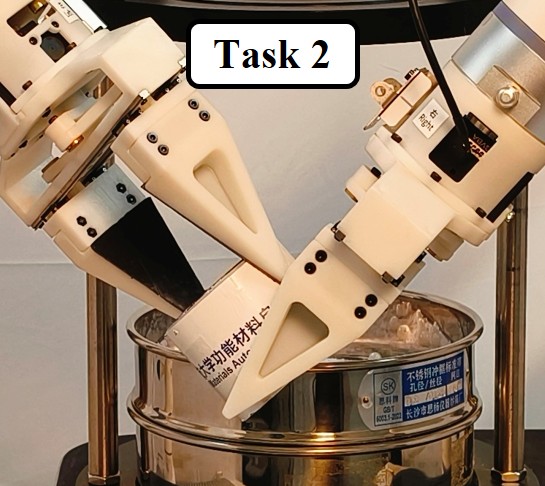}\label{fig1_b}} \\
\vspace{-0.30cm}
\subfloat[]
  {\includegraphics[width=0.24\textwidth]{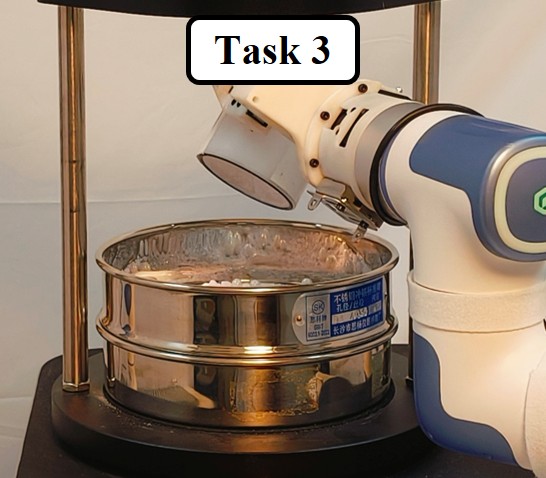}\label{fig1_c}} \hfill
\subfloat[]
  {\includegraphics[width=0.24\textwidth]{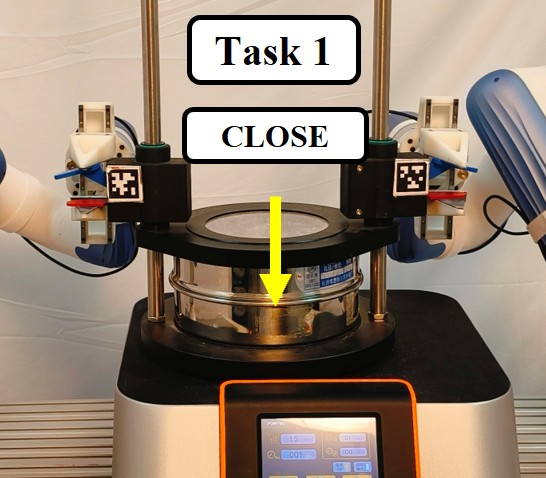}\label{fig1_d}} 
\caption{Key task sequences of dual-arm robotic operations: (a) Task 1, with yellow arrows indicating the opening direction; (b) Task 2, showing coordination and obstacle avoidance; (c) Task 3, following an optimized collision-free trajectory; (d) Task 1, with yellow arrows indicating the closing direction.}
\label{task}
\end{figure}

The goal of this study is to employ a dual-arm robot to operate a commonly used VSS in the laboratory for powder sample sieving operations. This operation task can be viewed as a composite workflow (All-Task), which is divided into three core manipulation tasks (Task 1-Task 3), as illustrated in Fig. \ref{task}:

\textbf{Task 1: Dual-arm Lid Manipulation} (Fig. \ref{fig1_a} and Fig. \ref{fig1_d}). The robot must precisely move its end-effector to the lid handle (clearance of 3 cm, which occupies 67.7\% of the gripper's maximum opening distance) for opening/closing, requiring joint-space planning with narrow-space constraints and collision avoidance.

\textbf{Task 2: Bimanual Handover} (Fig. \ref{fig1_b}). The left arm retrieves a powder sample container (PSC) and hands it over to the right arm. To improve execution efficiency, dual arm planning is carried out. This requires collision avoidance within narrow overlapping workspaces while maintaining the stability of the end effector.

\textbf{Task 3: Obstructed PSC Delivery} (Fig. \ref{fig1_c}). The right arm delivers the PSC into the VSS, ensuring collision-free trajectory planning with orientation constraints to prevent spillage.

\textbf{All-Task: Full VSS Operation Workflow}. The workflow follows the sequence: Task 1 (opening the VSS lid), Task 2 (PSC handover), Task 3 (delivering the PSC into the VSS), Task 3 (retrieving the PSC from the VSS), Task 2 (PSC return), and Task 1 (closing the VSS lid). The same task name is used for different processes (e.g., delivering and retrieving the PSC in Task 3) because these processes are symmetric in operation.


Existing methods face three key limitations in addressing these challenges:

\textbf{Inefficient Sampling in Narrow Passages}. Conventional sampling-based planners waste significant computational resources in narrow passages (e.g., Task 1's 3 cm clearance), leading to slow convergence and high collision-check overhead.

\textbf{Suboptimal Trajectory Quality}. Directly executed paths from sampling-based planners often exhibit jerky motions and excessive waypoints, making them unsuitable for precise tasks.

\textbf{Limited Adaptability}. Manual trajectory programming requires extensive effort, while learning-based approaches demand large amounts of training data, limiting their applicability.

To address these limitations, we propose a hierarchical planning framework that combines two key components: \textbf{Prior-Guided Path Planning} and \textbf{Multi-Step Trajectory Optimization}. Our approach directly targets both the core challenges of VSS operations and the specific needs of each task:

For inefficient sampling in narrow passages, we develop a \textbf{Prior-Guided Path Planning} strategy based on a finite Gaussian mixture model (FGMM) \cite{FGMM}. By leveraging prior knowledge of collision-free regions, this method significantly improves sampling efficiency in narrow passages, enabling faster exploration and convergence in constrained environments.

For suboptimal trajectory quality, we propose a \textbf{Multi-Step Trajectory Optimization} pipeline that combines path shortening, waypoint simplification, joint angle constraints, and B-spline smoothing. This ensures smooth and energy-efficient trajectories while maintaining pose stability.

For limited adaptability, our hierarchical planning framework eliminates the need for manual trajectory programming or extensive training data. By leveraging efficient sampling and optimization techniques, the framework can generate collision-free and executable trajectories in a short time, making it highly adaptable.

The effectiveness of our framework is validated through physical experiments on a dual-arm robotic platform. The system successfully completes the All-Task, demonstrating its practical applicability in real-world laboratory automation scenarios. Experimental results show significant improvements over baseline methods: planning time is
reduced by up to \textbf{80.4\%}, and waypoints are decreased by \textbf{89.4\%}. These results confirm that our approach not only addresses the core challenges of VSS operations but also provides a robust and efficient solution for complex laboratory automation tasks.

\begin{figure*}[t]
\centering
\includegraphics[width=0.9\textwidth]{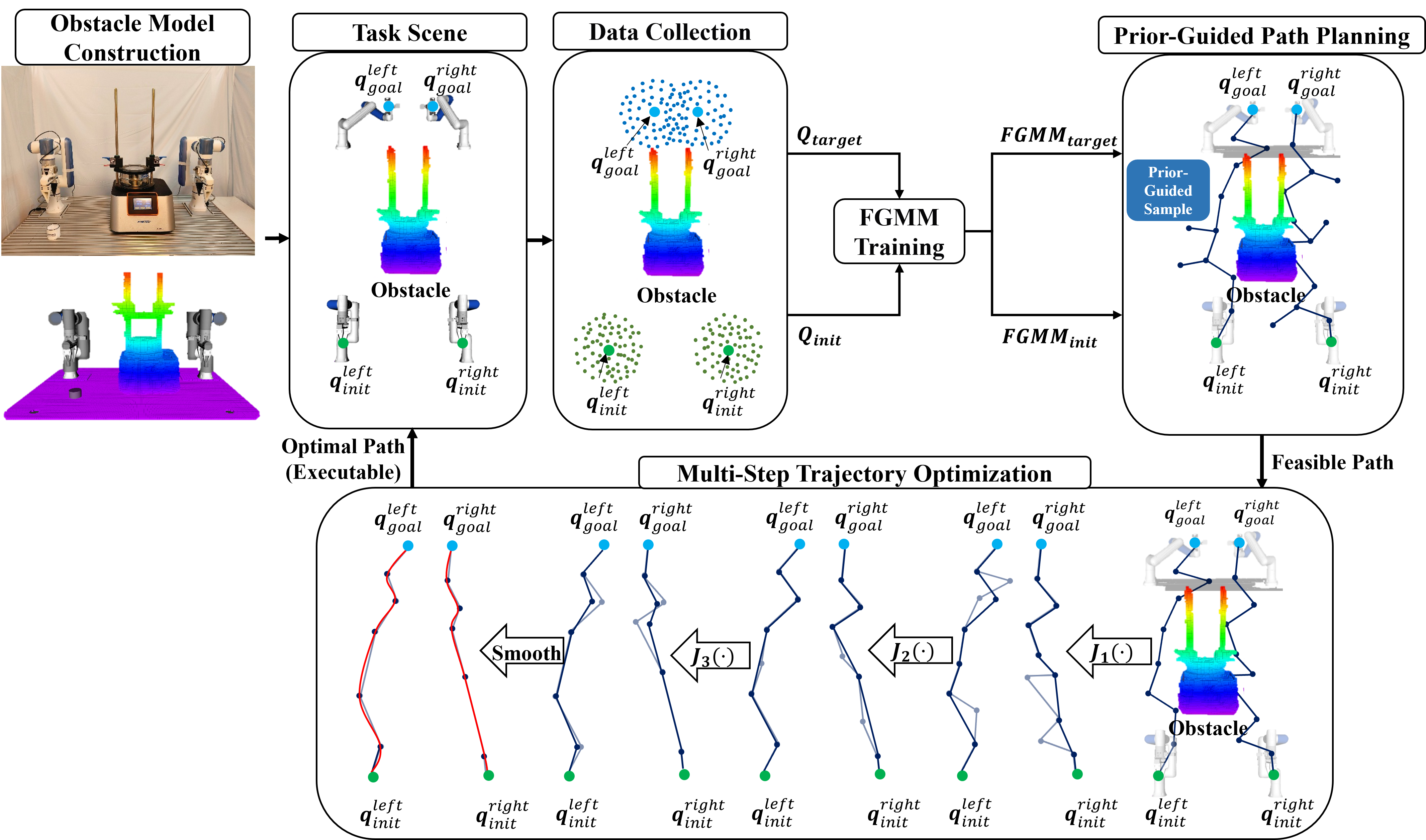}
\caption{System overview of the obstacle model construction and hierarchical planning framework.}
\label{methods}
\end{figure*}

\section{Related Works}
\subsection{Application of Dual-Arm Robots in Materials Laboratory Automation}
Dual-arm robots have been increasingly applied in materials laboratory automation. For instance, Heidi Fleischer et al. \cite{6} utilized an improved motion planning method to enable dual-arm robots to perform analytical measurements. Walker et al. \cite{8} automated viscosity estimation tasks, while Jiang et al. \cite{9} enabled dual-arm robots to mimic scientists' actions for material distribution. However, these studies often rely on predefined trajectories and assume obstacle-free environments, limiting their applicability in real-world scenarios where obstacle avoidance is critical. While some works, such as \cite{7}, have explored obstacle avoidance for dual-arm robots, they do not address the high-precision requirements of complex and constrained laboratory environments. Our work focuses on the challenge of high-precision obstacle avoidance in such settings, particularly when operating large experimental equipment, filling the research gap in this area.
\subsection{Motion Planning for Dual-Arm Robots}
Motion planning methods for dual-arm robots include graph search-based \cite{10,11,jiang2025social,jiang2024social}, optimization-based \cite{12,13,14,15}, and sampling-based approaches \cite{16,17,18,19}. Graph search algorithms like A* \cite{11} struggle in high-dimensional spaces due to exponential computational complexity. Optimization-based methods, while capable of incorporating desired trajectory properties, often depend on good initialization and can get trapped in local minima. Sampling-based methods, such as RRT and its variants (e.g., Goal-Bias RRT \cite{2}, RRT-Connect \cite{3}, PB-RRT* \cite{4}, and Neural RRT* \cite{5}), offer probabilistic completeness but face challenges in sampling efficiency and trajectory smoothness. While these methods have shown success in low-dimensional spaces, their application to high-dimensional dual-arm scenarios remains limited. Our work introduces target prior information to guide sampling and performs Multi-Step optimization on initial trajectories, ensuring efficient, safe, and smooth path planning for dual-arm robots.

\section{Method}
Our hierarchical planning framework addresses the challenges of efficient path planning and trajectory optimization for dual-arm robotic operations in narrow passages. It consists of two key components: Prior-Guided Path Planning and Multi-Step Trajectory Optimization. 

In the following subsections, we first describe obstacle model construction, then detail the Prior-Guided Path Planning method, and finally present the Multi-Step Trajectory Optimization pipeline.

\subsection{Obstacle Model Construction}
To enhance collision detection efficiency, obstacle models are constructed using minimum bounding boxes (MBBs) for regular-shaped objects, including the dual-arm robotic system links and the PSC. The two-finger gripper is decomposed into two independently moving components, each with its MBB, facilitating precise collision detection during manipulation and handover processes.

For complex industrial equipment, such as the VSS, lacking an accurate physical model and featuring intricate curved surfaces, Instant-NGP \cite{muller2022instant} is employed for reconstruction. The VSS is captured using hemispherical images and processed with COLMAP \cite{fisher2021colmap} for pose estimation, generating meshes within one minute. The final obstacle model, shown in Fig. \ref{methods} (left side), includes an OctoMap \cite{OctoMap} (resolution: 0.01m) representing the workspace, with the robotic arm and PSC depicted by MBBs. The gripper's MBB is scaled to accommodate the PSC's small size, ensuring collision-free operations.

\subsection{Prior-Guided Path Planning}
The traditional RRT-Connect algorithm relies on uniform random sampling, leading to slow convergence in complex constrained environments. To address this, we leverage prior knowledge of collision-free joint configurations near the start and goal points. Specifically, we model these regions using an FGMM, capturing the spatial distribution of valid configurations. The construction of the FGMM enables a more efficient sampling strategy, which is then integrated into the RRT-Connect framework to form the Prior-Guided RRT-Connect algorithm. In the following sections, we will detail the FGMM model, data collection process, FGMM training, selection of the number of Gaussian components, and the implementation of the Prior-Guided RRT-Connect algorithm.
\subsubsection{FGMM}
FGMM is a probabilistic model that assumes all data points are generated from a mixture of a finite number of Gaussian distributions with unknown parameters. FGMM can be utilized not only for clustering but also for probability density estimation. Furthermore, it is capable of generating new samples. The mathematical formulation of FGMM is as follows:
\begin{equation}\label{eq:ch3_04_px}
    p(\bm{q}| \bm{\theta} )=\sum\nolimits_{k=1}^{K}{{{\pi }_\text{k}}}\phi (\bm{q}|{{\bm{\theta} }_\text{k}})
,\end{equation}
where $\bm{q} \in \mathbb{R}^\text{n}$ represents the joint configuration of the dual-arm robot in this work, $K$ is the number of Gaussian components, and $\theta_\text{k} = \{\bm{\mu}_\text{k}, \bm{\Sigma}_\text{k}\}$ and $\bm{\theta} = \{\theta_\text{1}, \dots, \theta_\text{K}\} = \{\{\bm{\mu}_\text{1}, \bm{\Sigma}_\text{1}\}, \dots, \{\bm{\mu}_\text{K}, \bm{\Sigma}_\text{K}\}\}$ denote the local and global parameters of the Gaussian mixture model, respectively, with $\bm{\mu}_\text{k} \in \mathbb{R}^\text{n}$ being the mean vector and $\bm{\Sigma}_\text{k}\in \mathbb{R}^{\text{n} \times \text{n}}$ being the covariance matrix of the $k$-th component.

The term $\phi(\bm{q}|{\theta}_\text{k})$ represents the probability density function of the $k$-th Gaussian component, defined as:

\begin{equation}\label{eq:gaussian_pdf}
    \phi(\bm{q}|{\theta}_\text{k}) = \frac{1}{(2\pi)^{n/2}|\bm{\Sigma}_\text{k}|^{1/2}} {{e}^{-\frac{1}{2}{{{(\bm{q}-\bm{\mu}_\text{k} )}^{T}}{{\bm\Sigma}_\text{k}^{-1}}(\bm{q}-\bm{\mu}_\text{k} )}}}.
\end{equation}

The weights $\pi_\text{k}$ represent the mixing coefficients of the Gaussian components, satisfying the following constraints:
\begin{equation}\label{eq:pi_constraint}
    \sum_{k=1}^K \pi_\text{k} = 1.
\end{equation}

The unknown parameters in the FGMM are the mixing coefficients $\pi_k$ and the component parameters $\bm{\theta} = \{\bm{\mu}_\text{k}, \bm{\Sigma}_\text{k}\}_{k=1}^K$. In this work, these parameters are estimated from the data using the Expectation-Maximization (EM) algorithm \cite{EM}.
\subsubsection{Data Collection}
Considering that if narrow passages near the start and goal configurations significantly reduce sampling efficiency, collecting collision-free configurations around these regions provides more meaningful guidance compared to random or uniform sampling of joint configurations. To achieve this goal, we propose a target-oriented, variance-adaptive Gaussian sampling method centered around the target or starting points, which dynamically adjusts the sampling variance to construct a high-quality training dataset.

The process begins with the initialization of key parameters, including the dataset size $M$, the maximum standard deviation $\sigma_{\text{max}}$, and the initial standard deviation $\sigma = \sigma_{\text{init}}$. Given a target configuration $\bm{q}$, an independent $n$-dimensional Gaussian distribution is defined as:
\begin{equation}\label{eq:gaussian_pdf_}
    p(\bm{q}|\bm{\mu} ,\bm{\Sigma} ) = \frac{1}{(2\pi)^{n/2}|\bm{\Sigma}|^{1/2}} {{e}^{-\frac{1}{2}{{{(\bm{q}-\bm{\mu} )}^{T}}{{\bm\Sigma}^{-1}}(\bm{q}-\bm{\mu} )}}},
\end{equation}
where the covariance matrix $\bm{\Sigma}$ is diagonal and the variances of all $n$ joints are assumed to be identical, i.e.,  
\begin{equation}\label{eq:cov_matrix}
    \bm{\Sigma} = \sigma^2 \cdot \text{diag}(1, 1, \dots, 1),
\end{equation}  
with \( \sigma_\text{i} = \sigma \) for all \( i = 1, 2, \dots, n \). This assumption implies that each joint has the same variance, ensuring isotropic sampling behavior across the configuration space.

Next, a candidate configuration \( \bm{q}_\text{sample} \) is sampled from \( \mathcal{N}(\bm{\mu}, \bm{\Sigma}) \), where \( \bm{\mu} = \bm{q}_{\text{target}} \) and \( \sigma_\text{i} = \sigma \) for all \( i \). The validity of \( \bm{q} \) is then assessed through collision detection. If the sampled point is collision-free, it is added to the dataset \( \bm{Q} \); otherwise, it is discarded. To facilitate exploration in complex environments and escape from narrow passages, the standard deviation \( \sigma \) is adaptively updated based on the following rule:  
\begin{equation}\label{eq:sigma_update}
    \sigma = \begin{cases}
        1.1\sigma, & \sigma \leq \sigma_{\text{max}} \\
        \sigma_{\text{init}}, & \sigma > \sigma_{\text{max}}
    \end{cases}.
\end{equation}  
This process iterates until the dataset size reaches \( M \). The final dataset \( \bm{Q} \) consists of collision-free configurations surrounding the target, enabling the fitting of an FGMM model. This approach effectively increases the sampling probability in narrow passages (traps), thereby improving the robot's ability to navigate through constrained environments.

\subsubsection{FGMM Training}
 The training of the FGMM is the process of estimating the parameters, including the weights \( \pi_\text{k} \), the means \( \bm{\mu}_\text{k} \), and the covariance matrices \( \bm{\Sigma}_\text{k} \), based on the dataset \( \bm{Q} \). This is typically achieved using the EM algorithm.

Given the observed data points \( \bm{q}_\text{j} \in \bm{Q}, j=1,2,\dots,M \), the algorithm first initializes the parameters. The number of Gaussian components is set to \( K \), and initial parameters \( \bm\theta^{(0)} \) are estimated using the K-Means clustering algorithm. After initialization, the algorithm iterates as follows:

\begin{itemize}
    \item \textbf{E-Step}: Compute the expectation of the conditional probability of the latent variables, i.e., the probability that the observed data point \( \bm{q}_\text{j} \) belongs to the \( k \)-th Gaussian component given the current parameter estimates \( \theta_\text{k} \):
\begin{equation}\label{eq:ch3_06_pk}
    P(k|\bm{q}_\text{j}) = \frac{\pi_\text{k} \phi (\bm{q}_\text{j} | {\theta}_\text{k})}
    {\sum\nolimits_{k=1}^{K} \pi_\text{k} \phi (\bm{q}_\text{j} | {\theta}_\text{k})}, \quad j = 1,2,\dots,M.
\end{equation}

    \item \textbf{M-Step}: Maximize the expectation by updating the parameters, including the weights \( \pi_k \), the means \( \bm{\mu}_k \), and the covariance matrices \(\bm{\Sigma}_k \), to maximize the probability of the observed data \( \bm{Q} \). The total responsibility for the \( k \)-th component is given by \( M_k = \sum\nolimits_{j=1}^{M} P(k|x_j) \), which represents the effective number of observations assigned to the \( k \)-th Gaussian component. The updated parameters are computed as follows:

\begin{equation}\label{eq:ch3_07_pi}
    \pi_\text{k} = \frac{M_\text{k}}{M}, \\[8pt]
\end{equation}
\begin{equation}\label{eq:ch3_07_pi_}
    \bm{\mu}_\text{k} = \frac{1}{M_\text{k}} \sum\nolimits_{j=1}^{M} P(k|\bm{q}_\text{j}) \bm{q}_\text{j}, \\[8pt]
\end{equation}
\begin{equation}\label{eq:ch3_07_sigma}
    \bm{\Sigma}_\text{k} = \frac{1}{M_\text{k}} \sum\nolimits_{j=1}^{M} P(k|\bm{q}_\text{j}) (\bm{q}_\text{j} -\bm{\mu}_\text{k})(\bm{q}_\text{j} - \bm{\mu}_\text{k})^\top.
\end{equation}
\item \textbf{Convergence Check}: The log-likelihood function is computed as follows:
\begin{equation}\label{eq:ch3_H}
    H(\bm{\theta}) = \ln L(\bm{\theta}) = \sum\nolimits_{j=1}^{M} \ln p(\bm{q}_\text{j} | \bm{\theta}),
\end{equation}
where \( p(\bm{q}_\text{j} | \bm{\theta}) \) represents the probability density of the observed data point \( \bm{q}_\text{j} \) under the FGMM. The iteration continues until the log-likelihood function converges to a stable value. If it has not yet converged, the algorithm returns to the E-step and continues the process.

\end{itemize}

After the iterations are complete, the algorithm outputs the trained FGMM model, including the estimated weights \( \pi_\text{k} \), mean vectors \( \bm{\mu}_\text{k} \), and covariance matrices \( \bm{\Sigma}_\text{k} \).

\begin{algorithm}[t]
	\caption{Prior-Guided RRT-Connect($\bm{q}_\text{init}$, $\bm{q}_\text{goal}$)}
	\label{alg:improved RRT_Connect} 
	\begin{algorithmic}[1] 
		\STATE $\textbf{Initialize}: StepSize, P_{\text{bias} }, T_{0}.\text{init}(\bm{q}_{\text{init} } ), T_{1} .\text{init}(\bm{q}_{\text{goal}} )$
		\STATE $X_{\text{init} }  \gets  \text{GetExemplars}(\bm{q}_{\text{init} } )$
        
        \STATE $X_{\text{goal} }  \gets  \text{GetExemplars}(\bm{q}_{\text{goal} } )  $
		\STATE ${FGMM\text{-}Init}  \gets \text{BuildFGMM}(X_{\text{init} })$
        \STATE ${FGMM\text{-}Goal}  \gets \text{BuildFGMM}(X_{\text{goal} })$
		\STATE $D \gets \text{Distance}(\bm{q}_{\text{init} }, \bm{q}_{\text{goal}})$
		\STATE $TowardGoal   \gets \mathrm{True} $
		\FOR{$i \gets 0$ to $N$}
			\STATE $p \gets \text{Random} (0, 1)$
			\IF{$ p < P_{\text{bias } } $}
				\IF{$TowardGoal$}
					\STATE $\bm{q}_{\text{sample} }  \gets \text{PriorInfoSampling} ({FGMM\text{-}Goal})$
				\ELSE
					\STATE $\bm{q}_{\text{sample} }  \gets \text{PriorInfoSampling} ({FGMM\text{-}Init})$
				\ENDIF
			\ELSE
				\STATE $\bm{q}_{\text{sample} }  \gets \text{CurrentInfoSampling} (T_{0}, D, \bm{q}_{\text{init} }, \bm{q}_{\text{goal}})$
			\ENDIF
			\STATE $\bm{q}_{\text{near} }  \gets \text{NearestNode}(T_{0}, \bm{q}_{\text{sample}})$
			\STATE $\bm{q}_{\text{a\_reach} }  \gets \text{Extend}(T_{0}, \bm{q}_{\text{near}}, \bm{q}_{\text{sample}})$
			\STATE $\bm{q}_{\text{near} }  \gets \text{NearestNode}(T_{1},\bm{q}_{\text{a\_reach}})$
			\STATE $\bm{q}_{\text{b\_reach} }  \gets \text{Extend}(T_{1}, \bm{q}_{\text{near}}, \bm{q}_{\text{a\_reach}})$
			\STATE $\bm{q}_\text{init}$, $\bm{q}_\text{goal} \gets \bm{q}_\text{goal}$, $\bm{q}_\text{init} $	
			\STATE $T_{0}, T_{1} \gets T_{1}, T_{0}$
			\STATE $TowardGoal   \gets \mathrm{False} $
			\IF{$\bm{q}_{\text{a\_reach}}=\bm{q}_{\text{b\_reach}}$}
				\RETURN $T_{0}, T_{1}$
			\ENDIF
		\ENDFOR
	\end{algorithmic}
\end{algorithm}

\subsubsection{Prior-Guided RRT-Connect}

Building upon the FGMM-based environmental modeling around critical nodes, we propose a Prior-Guided RRT-Connect algorithm that synergizes offline prior information with online tree growth dynamics. By probabilistically biasing sampling toward FGMM-derived high-likelihood regions and adaptively adjusting exploration directions using real-time tree state feedback, our method achieves accelerated convergence to feasible paths in cluttered environments while maintaining probabilistic completeness.  

The algorithm alternates between expanding two trees (\( T_0 \) and \( T_1 \)) toward each other. For each tree, the sampling process combines prior information (from FGMM) and current information (from the tree's growth state) based on a probability threshold \( P_{\text{bias}} \). If a random probability \( p \) is less than \( P_{\text{bias}} \), the algorithm samples based on prior information; otherwise, it samples based on the current state of the tree.

\paragraph{Prior-Guided Sampling}
The Prior-Guided sampling method utilizes the FGMM, which is built around the start and goal points. This method biases the sampling process toward regions with high probability, ensuring that the search focuses on more promising areas of the configuration space. The mathematical formulation for prior information sampling is given by:
\begin{equation}\label{eq:prior_info_sampling}
    \bm{q}_{\text{sample}} \sim \sum_{k=1}^{K} \pi_k \mathcal{N}(\bm{\mu}_k, \bm{\Sigma}_k).
\end{equation}

\paragraph{Current Information Sampling}
The current information sampling dynamically adjusts the sampling based on the tree's growth state. It leverages real-time information to guide the tree toward the target configuration. The sampling formula is defined as:
\begin{equation}\label{eq:current_info_sampling}
    \bm{q}_{\text{sample}} \sim \mathcal{N}(\bm{q}_{\mu}, 1 - \text{rate}),
\end{equation}
where \( \bm{q}_{\mu} \) is the mean configuration calculated as:
\begin{equation}\label{eq:mean_q_mu}
    \bm{q}_{\mu} = \bm{q}_{\text{init}} + \text{rate}(\bm{q}_{\text{goal}} - \bm{q}_{\text{init}}),
\end{equation}
and the rate is determined by the distance between the current node and the goal:
\begin{equation}\label{eq:rate_calculation}
    \text{rate} = \frac{D - d}{D},
\end{equation}
where \( d \) is the distance from the current tree node to the goal, and \( D \) is the total distance from the initial configuration to the goal.

\paragraph{Tree Expansion and Collision Handling}
During tree expansion, the algorithm alternates between \( T_0 \) and \( T_1 \). If a collision is detected, the current tree stops expanding, and the other tree continues. The collision detection mechanism ensures a safe distance \( d_{\text{safe}} \) between the robot and obstacles, as well as between the robot's links. Specifically, the following conditions are enforced:

\begin{equation}\label{eq:ch3_12_collision_env}
    \|\bm{L}_i - \bm{O}_{\text{obs}}\|_2 \geq d_{\text{safe}}, \quad \forall i = 1, 2, \dots, N_{\text{link}},
\end{equation}
\begin{equation}\label{eq:ch3_12_collision_self}
    \|\bm{L}_i - \bm{L}_j\|_2 \geq d_{\text{safe}}, \quad \forall i, j = 1, 2, \dots, N_{\text{link}}, \quad i \neq j,
\end{equation}
where \( \bm{L}_i \) is the inflated bounding box of the \( i \)-th link, \( \bm{O}_{\text{obs}} \) represents obstacles (modeled as an OctoMap) and \( N_{\text{link}} \) is the total number of robot links.

The overall algorithm is shown in Algorithm \ref{alg:improved RRT_Connect}, and the key procedures are as follows:
\begin{itemize}
    \item \textbf{GetExemplars}($\bm{q}$): A target-oriented, variance-adaptive data collection method that dynamically adjusts sampling variance around the start or goal configurations.
    \item \textbf{BuildGMM}($\bm{Q}$): Constructs a $FGMM$ using the exemplar dataset \( \bm{Q} \) to model task-relevant prior information.
    \item \textbf{Distance}($\bm{q}_1, \bm{q}_2$): Computes the Manhattan distance between two joint configurations \( \bm{q}_1 \) and \( \bm{q}_2 \).
    \item \textbf{PriorInfoSampling}($FGMM$): Samples from the $FGMM$ to guide the search toward high-probability regions.
    \item \textbf{CurrentInfoSampling}($T, D, \bm{q}_{\text{init}}, \bm{q}_{\text{goal}}$): Dynamically adjusts sampling based on the tree's current state to guide it toward the goal.
    \item \textbf{NearestNode}($T, \bm{q}$): Finds the nearest node to \( \bm{q} \) in the tree \( T \).
    \item \textbf{Extend}($T, \bm{q}_{\text{near}}, \bm{q}_{\text{sample}}$): Extends from \( \bm{q}_{\text{near}} \) toward \( \bm{q}_{\text{sample}} \). The extension terminates if a collision is detected or if the distance to \( \bm{q}_{\text{sample}} \) begins to increase. The function returns \( \bm{q}_{\text{new}} \), which is either \( \bm{q}_{\text{sample}} \) or the last collision-free node along the extension path.
\end{itemize}

\subsection{Multi-Step Trajectory Optimization}

\begin{algorithm}[t]
	\caption{Multi-Step Trajectory Optimization($\bm{\tau}$)}
	\label{alg:Tra_opt} 
	\begin{algorithmic}[1] 
		\STATE $\textbf{Initialize}: D_\text{thre}, M, \alpha _\text{max}$
		\FOR{$i \gets 0$ to $M$} 
			\STATE $\bm{q}_{i}, \bm{q}_{j} \gets \text{RandomSample} (\bm{\tau})$
			\STATE $T.\text{init}()$
			\STATE $\bm{q}_\text{reach} \gets \text{Extend}(T,\bm{q}_{i}, \bm{q}_{j})$
			\IF{ $\bm{q}_\text{reach} \ne \bm{q}_{j}$}
				\STATE \textbf{continue}
			\ENDIF
			\IF{$\text{NewPathLen}(T, \bm{q}_{i}, \bm{q}_{j}) < \text{PathLen}( \bm{q}_{i}, \bm{q}_{j}) $}
				\STATE $\bm{\tau} \gets \text{Update}(T,\bm{q}_{i}, \bm{q}_{j})$
			\ENDIF
		\ENDFOR
		\STATE $\bm{\tau} \gets \text{DouglasPeuckerAlgo}(\bm{\tau},D_\text{thre})$
		\FOR{each node $\bm{q}_{i}\in P$}
			\FOR{each joint ${q}_{i,j}\in \bm{q}_{i}$}
				\IF{${G}({q}_{i,j}) \ge \alpha _\text{max}$}
					\STATE $\text{RefineJoint}({q}_{i,j})$
				\ENDIF
			\ENDFOR
		\ENDFOR
		\STATE $\bm{\tau} \gets \text{CubicBspline}(\bm{\tau})$
		\RETURN $\bm{\tau}$
	\end{algorithmic}
\end{algorithm}

To refine the raw path into a suitable trajectory for execution, we propose a Multi-Step trajectory optimization pipeline. The optimization process minimizes the total path length, simplifies the waypoint sequence, and ensures smooth joint rotations. The final optimized trajectory $\bm{\tau}^*$ is constructed using cubic B-spline interpolation, providing a smooth and continuous path suitable for robotic motion.

\subsubsection{Path Length Minimization}

The first step in the optimization process is minimizing the path length. Given that the planning algorithm may generate suboptimal paths, we aim to reduce the total length by performing \textit{restart node optimization}. This method leverages a cost function

\begin{footnotesize}
\begin{equation}\label{J1}
J_1(\bm{\tau}) \!= \!\arg\!\min_{\!\bm{\tau} \in \bm{C}_{\text{free}}} \!\!\!\left\{
  \!\sum_{i=1}^{\operatorname{Size}(\bm{\tau})}\!\! \left\| \bm{q}_i \!- \!\bm{q}_{i-1} \right\|_1 
  \middle| \!\!\ \bm{\tau}(0) \!= \!\bm{q}_{\text{init}},\bm{\tau}(1) \!= \!\bm{q}_{\text{goal}} \!\!\right\},
\end{equation}
\end{footnotesize}which minimizes the total path length while ensuring that the path $\bm{\tau}$ remains feasible within the free configuration space \( \bm{C}_{\text{free}} \).

During restart node optimization, two random waypoints are selected from the existing path. A new tree is initialized between these waypoints, and the algorithm attempts to find a collision-free path between them. If the new path is shorter than the original one, the waypoints are replaced. This process is repeated for \( M \) iterations, progressively shortening the overall trajectory. The restart node optimization procedure is outlined in lines 2--12 of Algorithm \ref{alg:Tra_opt}.

\subsubsection{Waypoints Simplification}
Although the restart node optimization step reduces the path length, the trajectory $\bm{\tau}$ may still contain many redundant waypoints, which can lead to frequent minor direction changes or minute position adjustments. To simplify the trajectory while preserving the essential characteristics of the path, we formulate the optimization objective \( J_2(\bm{\tau}) \) (Eq. \ref{J2}) and apply the Douglas-Peucker algorithm \cite{Douglas-Peucke}. The algorithm recursively simplifies the path by connecting the start and end points of a segment to form a reference line. If the maximum distance \( d_{\text{max}} \) between the intermediate waypoints and the reference line is smaller than a threshold \( D_{\text{thre}} \), the intermediate points are discarded. Otherwise, the point corresponding to \( d_{\text{max}} \) is retained, and the path is divided into two sub-segments for further processing. The threshold \( D_{\text{thre}} \) controls the simplification intensity. In our implementation, we set \( D_{\text{thre}} = 0.005 \, \text{m} \), meaning that waypoints with deviations smaller than 5 mm are removed, effectively simplifying the path while preserving its essential features.

\begin{equation}\label{J2}
J_2(\bm{\tau}) = \operatorname*{argmin}_{\bm{\tau} \in \bm{C}_{\text{free}}} \left\{ \operatorname{Size}(\bm{\tau}) \ \middle| \ D_{\text{thre}} \right\}
\end{equation}

\subsubsection{Joint Rotation Constraints}
One critical aspect of trajectory optimization is ensuring that the robotic arm moves smoothly between waypoints, without sharp or abrupt joint rotations. To address this, we introduce a joint rotation angle constraint \( J_3(\bm{\tau}) \), which limits the maximum allowed joint angle change per unit distance (rad/m). The joint rotation constraint is defined as follows:

\begin{equation}
\alpha_{i,j} = \frac{|q_{i,j}-q_{j,i-1}|}{\|\bm{p}_i-\bm{p}_{i-1}\|_2},
\end{equation}

\begin{equation}\label{J3}
\begin{aligned}
{G}(q_{i,j}) = 
\begin{cases} 
|\alpha_{i,j}| + |\alpha_{i+1,j}| & \text{if } \alpha_{i,j} \alpha_{i+1,j} < 0 \\ 
0 & \text{if } \alpha_{i,j} \alpha_{i+1,j} \geq 0 
\end{cases}, 
\end{aligned}
\end{equation}

\begin{footnotesize}  
\begin{equation}\label{J4}
J_3(\bm{\tau}) = \operatorname*{argmin}_{\bm{\tau} \in \bm{C}_{\text{free}}} \left\{ 
\sum_i^{Size(\bm{\tau})} \sum_j^n {G}(q_{i,j}) \ \middle| \ {G}(q_{i,j}) < \alpha_{\max} \right\},
\end{equation}
\end{footnotesize}where \( \alpha_{i,j} \) measures the angular change of joint \( j \) between waypoints \( i \) and \( i-1 \), \( q_{i,j} \) is the joint angle, and \( \bm{p}_i \) is the position of the Tool Center Point (TCP) associated with the arm containing joint \( j \). The discontinuity metric ${G}(q_{i,j})$ quantifies abrupt joint reversals only when the angular changes between consecutive waypoints have opposite signs. In our implementation, the maximum allowable angular velocity is set to \( \alpha_{\max} = \pi \, \text{rad/m} \), meaning each joint can rotate at most one full circle per meter of TCP movement.
\subsubsection{Cubic B-Spline Interpolation}

After performing path shortening, waypoint simplification, and enforcing joint rotation constraints, the final step is to generate a smooth and continuous trajectory that the robotic arm can execute. For this, we use cubic B-spline interpolation. The B-spline method provides a smooth curve that passes through the initial and goal configurations, ensuring the trajectory is feasible and continuous.

\subsubsection{Summary of the Optimization Process}

The Multi-Step Trajectory Optimization algorithm is summarized in Algorithm \ref{alg:Tra_opt}. By combining restart node optimization, waypoint simplification, joint rotation constraints, and cubic B-spline interpolation, we achieve a smooth, efficient, and safe trajectory for the robotic arm. This optimized trajectory not only minimizes path length and energy consumption but also ensures that the arm moves safely and smoothly through the configuration space.

\section{Experiment}
\begin{figure*}[t]
\centering
\includegraphics[width=0.98\textwidth]{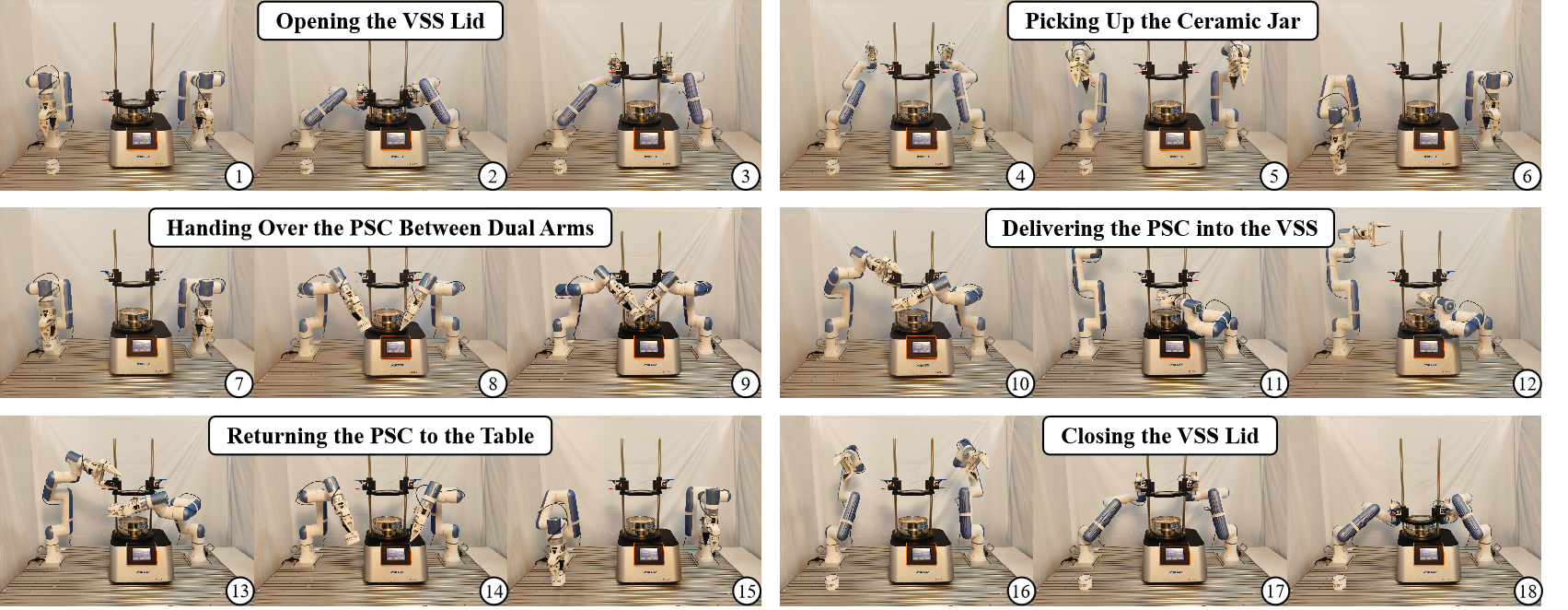}
\caption{The integrated experiment of dual-arm VSS manipulation.}
\label{sac-arm}
\end{figure*}
The experimental setup utilizes a DOBOTX-Trainer Al Robot Operation Platform, featuring NOBOT Nova2 dual-arm robots and built-in grippers. The POWTEQ VSS SS2000 sifting machine (47 cm × 47 cm × 87 cm) features a cylindrical delivery region (20 cm diameter, 15 cm height) located at a height of 34 cm. The platform measures 140 cm × 100 cm, with cylindrical PSC containers (7.8 cm diameter, 6.1 cm height). The robotic arms are spaced 108 cm apart, with a 62.5 cm workspace radius and 20 cm grippers. Algorithms run on a workstation with an AMD Ryzen Threadripper PRO 3975WX 32-core processor and an NVIDIA RTX A6000 GPU.

\subsection{Prior-Guided Sampling Results}
To validate that the proposed sampling method accelerates the sampling process and reduces planning time, we compared it with random sampling and goal-bias sampling methods. For the proposed Prior-Guided sampling method, the parameters for collision-free configuration sampling were set with the variance parameters \(\sigma_{\text{min}}\) and \(\sigma_{\text{max}}\) at \(0.0872\) and \(0.3491\) respectively, the number of Gaussian components \(K\) at \(2\), and the size of the collision-free configuration dataset \(M\) at \(500\). 
 
 The experimental results are presented in Table~\ref{tab:sample}. For each of the three tasks, 10 trials were conducted, and the average number of extended nodes (rounded up) and average planning time were recorded. The proposed method consistently outperforms both baselines, reducing the average extended nodes by 71.4\%, 30.9\%, and 64.3\% for Tasks 1--3 compared to random sampling, and by \textbf{78.0\%}, 47.7\%, and 79.0\% compared to goal-bias sampling, respectively. The average planning time is also significantly reduced, with Task 1 decreasing by 73.2\% and \textbf{80.4\%}, Task 2 by 30.7\% and 10.5\%, and Task 3 by 52.2\% and 72.1\% compared to the two baselines. These results demonstrate the method's effectiveness in improving planning efficiency across all three tasks, highlighting its ability to reduce computational overhead and enhance trajectory generation performance.

\subsection{Multi-Step Trajectory Optimization Results}
\begin{table}[t]
\caption{The comparison results of different sampling methods based on RRT-Connect algorithm}
\resizebox{0.48\textwidth}{!}{%
\label{tab:sample}
\begin{tabular}{cccc}
\hline
Stage                  & Sampling Method    & Extended Nodes (\#) & Time (s)        \\ \hline
\multirow{3}{*}{Task 1} & Random Sampling    & 7793             & 709.73          \\
                       & Goal-bias Sampling & 10139            & 972.72          \\
                       & \textbf{Ours}               & \textbf{2227}    & \textbf{190.22} \\
\multirow{3}{*}{Task 2} & Random Sampling    & 162             & 13.47           \\
                       & Goal-bias Sampling & 214                & 10.43           \\
                       & \textbf{Ours}               & \textbf{112}     & \textbf{9.34}   \\
\multirow{3}{*}{Task 3} & Random Sampling    & 465              & 41.13           \\
                       & Goal-bias Sampling & 791              & 70.58           \\
                       & \textbf{Ours}               & \textbf{166}     & \textbf{19.67}  \\ \hline
\end{tabular}
}
\end{table}

\begin{table}[t]
\caption{The statistical results of the optimization algorithm}
\label{optimize}
\resizebox{0.48\textwidth}{!}{%
\begin{tabular}{ccccc}
\hline
Steps                                       & Metrics               & Task 1   & Task 2 & Task 3  \\ \hline
\multirow{3}{*}{Path Planning}              & Extended Nodes (\#)   & 2227  & 111  & 165   \\
                                            & Path Nodes (\#)       & 214   & 73   & 148   \\
                                            & Path Length (rad)     & 31.62 & 8.54 & 17.49 \\
\multirow{2}{*}{Path Length Minimization}   & Path Nodes (\#)       & 147   & 50   & 92    \\
                                            & Path Length (rad)     & 16.48 & 5.64 & 10.41 \\
Waypoints Simplification                    & Path Nodes (\#)       & \textbf{17}    & \textbf{7}    & \textbf{9}     \\
Joint Rotation Constraints                  & Refine Joints (\#)    & 3     & 2    & 1     \\ \hline
\end{tabular}
}
\end{table}
The results in Table~\ref{optimize} demonstrate the effectiveness of the proposed Multi-Step Trajectory Optimization pipeline, which includes path planning, path length minimization, waypoint simplification, and joint rotation constraints. The experiments aim to validate that each optimization step contributes to improving path quality. In the initial planning stage, feasible paths are generated, with Task 1 resulting in 214 path nodes and a path length of 31.62 radians, Task 2 achieving 73 nodes and 8.54 radians, and Task 3 achieving 148 nodes and 17.49 radians. The path length minimization process significantly reduces path nodes and length, decreasing Task 1 nodes by 31.1\% (to 147) and path length by \textbf{47.9\%} (to 16.48 rad), Task 2 nodes by 31.5\% (to 50) and path length by 34.0\% (to 5.64 rad), and Task 3 nodes by \textbf{37.8\%} (to 92) and path length by 40.5\% (to 10.41 rad). The waypoints simplification process further simplifies paths, reducing Task 1 nodes by 88.2\% (to 17), Task 2 nodes by 86.0\% (to 7), and Task 3 nodes by \textbf{89.4\%} (to 9). Finally, the joint rotation constraints refine joint angles, ensuring smooth trajectories. Specifically, Task 1 requires three joint adjustments, Task 2 requires two, and Task 3 requires one. Overall, the pipeline effectively reduces path complexity and length while maintaining smooth and executable trajectories for dual-arm robot operations across all three tasks.

\subsection{Integrated Experimental Results}
To validate the practical applicability of the proposed framework, we conducted the All-Task experiment, which involves executing the full multi-step workflow on the physical dual-arm robotic system. The trajectory of each subtask was regenerated and combined into a complete workflow, demonstrating the system's ability to operate the VSS in real-world conditions. The experiment completed the entire workflow, as illustrated in Fig.~\ref{sac-arm}, \textbf{and the corresponding robot videos are provided in the supplementary material}. This result confirms that the generated trajectories are executable on the physical robot, effectively validating the practicality of the framework for laboratory automation tasks.

\section{Conclusion}
This study presents a hierarchical planning framework for dual-arm robotic automation of VSS operations in materials laboratories. The framework addresses three critical challenges: inefficient sampling in narrow passages, suboptimal trajectory quality, and limited adaptability of existing methods. By integrating Prior-Guided Path Planning and Multi-Step Trajectory Optimization, our approach achieves significant improvements:

\begin{itemize}
\item \textbf{Efficient Planning}: The Prior-Guided sampling reduces extended nodes by up to 78.0\% and planning time by 80.4\% compared to goal-bias sampling, particularly in constrained tasks like Task 1 (3 cm clearance), where planning time drops from 972.72 seconds to 190.22 seconds.
\item \textbf{High-Quality Trajectories}: The optimization pipeline reduces path nodes by 89.4\% (e.g., Task 3) and shortens path lengths by 47.9\% (Task 1), ensuring smooth, collision-free motions critical for spillage prevention and stability.
\item \textbf{Practical Validation}: The framework successfully executes the All-Task on a physical dual-arm platform, confirming its real-world applicability.
\end{itemize}
These results demonstrate the framework's ability to overcome limitations in narrow-space manipulation, bimanual coordination, and trajectory smoothness, offering a robust solution for laboratory automation. Future work will integrate visual perception and extend the framework to dynamic environments while further expanding it to other material laboratory automation tasks.

\addtolength{\textheight}{-12cm}   






\bibliographystyle{IEEEtran}
\bibliography{IEEEabrv,mylib}

\end{document}